\documentclass[10pt, a4paper]{article}
\usepackage{lrec-coling2024} 

\usepackage{booktabs}
\usepackage{multirow}
\usepackage{tipa}

\newcommand{\pinjie}[1]{\textcolor{black}{#1}}

\title{Modeling Orthographic Variation Improves \\ NLP Performance for Nigerian Pidgin}

\name{Pin-Jie Lin$^{\spadesuit}$, Merel Scholman$^{\spadesuit \clubsuit}$, Muhammed Saeed$^{\spadesuit}$ and Vera Demberg$^{\spadesuit}$} 

\address{$^{\spadesuit}$ Language Science and Technology, Saarland University, Germany\\
$^{\clubsuit}$ Institute for Language Sciences, Utrecht University, the Netherlands \\
         \{pinjie, m.c.j.scholman, musaeed, vera\}@coli.uni-saarland.de}

\abstract{
Nigerian Pidgin is an English-derived contact language and is traditionally an oral language, spoken by approximately 100 million people. No orthographic standard has yet been adopted, and thus the few available Pidgin datasets that exist are characterised by noise in the form of orthographic variations. This contributes to under-performance of models in critical NLP tasks.
The current work is the first to describe various types of orthographic variations commonly found in Nigerian Pidgin texts, and model this orthographic variation.
The variations identified in the dataset form the basis of a phonetic-theoretic framework for word editing, which is used to generate orthographic variations to augment training data. We test the effect of this data augmentation on two critical NLP tasks: machine translation and sentiment analysis.
The proposed variation generation framework augments the training data with new orthographic variants which are relevant for the test set but did not occur in the training set originally. Our results demonstrate the positive effect of augmenting the training data with a combination of real texts from other corpora as well as synthesized orthographic variation, resulting in performance improvements of 2.1 points in sentiment analysis and 1.4 BLEU points in translation to English.
 \\ \newline  }

\begin{document}

\maketitleabstract



\section{Introduction}

Models developed for a variety of NLP tasks can give high-quality performance for resource-rich languages, such as English and French. However, for under-resourced languages such as many African and South-East Asian languages, NLP models show poor performance due to a lack of high-quality, sufficiently sized, publicly available datasets.\footnote{The Masakhane community aims to address this by strengthening and spurring NLP research in African languages; see \url{https://www.masakhane.io}.} 
In such cases, models might be particularly sensitive to noise in the data. 
One source of ``noise'' is orthographic variation -- that is, variations in the spelling of words in the data. Orthographic variation can be detrimental to the performance of NLP models, which are typically trained on curated datasets and tend to break when faced with noisy data \citep{bergmanis2020robust}.
The issue of orthographic variation is especially present for languages that do not have a standardized and normalized orthography yet, which is the case for many creoles and pidgins. 


The current work addresses the orthographic variation in one such language, namely Nigerian Pidgin.
This language has 100 million speakers, but is still largely absent from NLP research.
Nigerian Pidgin is a predominantly spoken language, without a normalized orthography in place \cite{mensah2021towards}. Consequently, written Nigerian Pidgin texts are characterized by a large proportion of orthographic variations.
These diverse orthographic variations contribute to a significant under-performance in critical tasks, such as sentiment analysis and machine translation.



We address this by synthesizing the orthographic variation at the phonological level, and subsequently training language models on data augmented with these variations.
Our contributions are as follows:
\begin{itemize}
    \item We are the first to provide an analysis of the various types of orthographic variations that occur in a variety of Nigerian Pidgin texts (the Bible, magazine texts and transcriptions of spoken conversations). These orthographic variations are found to be of a phonetic nature. 
    \item We propose a phonetic-theoretic framework for word editing, which can be used to generate orthographic variations to augment training data for language models. 
    \item We show performance improvement on two NLP tasks (machine translation and sentiment analysis) when including the augmented training data.
\end{itemize}

\section{Related work}

\paragraph{NLP research on Nigerian Pidgin}
Pidgins and creole languages have various unique features (e.g., grammar, morphology, lexicon) that make them interesting to study for various linguistic topics. Nevertheless, linguistic research and resources on these languages are limited \cite{lent2021language}. Nigerian Pidgin is no exception to this, although it has received more attention in recent years.
A few datasets now exist for Nigerian Pidgin, focusing on newspaper text \cite{ogueji2019pidginunmt,ndubuisi2019}, text from several magazines written by a religious society \cite{agic2019}, and task-specific data for named entity recognition \cite{oyewusi2021,adelani-etal-2021-masakhaner}, sentiment analysis \cite{oyewusi2020}, speech recognition \cite{ajisafe2020towards}, and transcribed spoken data \citep{caron2019surface}.

\citet{lin2023low} enriched the existing available parallel and monolingual Pidgin datasets to generate a high-quality fully parallel corpus of Nigerian Pidgin text across ten resources and five domains. 

\paragraph{Writing without a standard orthography}
An orthography is a set of rules or conventions that is used to represent language in a standardized system of writing.
When a language does not have a commonly used orthography, writers make decisions about orthography based on various criteria. The most dominant criterion is phonology: writers try to match the pronunciation of the word in the writing system, given (language-specific) assumptions about grapheme-to-phoneme mapping \citep{eskander2013processing}. Often, these assumptions come from other languages known to the writer; in the case of Nigerian Pidgin, this might be English or local Nigerian languages. 

\citet{deuber2007dynamics} present an analysis of orthographic choices in Nigerian Pidgin computer mediated communication. They report that for Pidgin lexical items that have English origins and mean the same as the origin, the English spelling is commonly adopted for Pidgin. In such cases, non-English spellings are used mainly for the symbolic purpose of indicating distance from English. For example, English `thing' is often written as Pidgin \textit{thing}, although a non-English spelling (\textit{tin}) is also possible. For Pidgin lexical items which have roots in English, but have developed distinct meanings in Pidgin, writers are more likely to adopt non-English spellings. For example, `done' has adopted a new meaning in Pidgin and is thus usually written as \textit{don}. For Pidgin lexical items of non-English origin (e.g., \textit{pikin} meaning `child'), writers tend to adopt the phonemic orthographies of local African languages.

\paragraph{Modeling orthographic variation}
%
\citet{eskander2013processing} address the orthographic variation found in Egyptian Arabic by normalizing the input data (i.e., transforming variations into a conventionalized orthography). Their results showed an error reduction of 69.2\% over a baseline on an unseen test set. While the approach seems promising, a normalization approach is not efficient for languages that lack a standardized orthography since (i) it is not always clear which word form is the standard one and should be taken as the base word, and (ii) new and unknown variations will likely be encountered in new datasets, which makes the process not scaleable. Rather, we propose to train a model to be able to deal with orthographic variation directly, without the preprocessing step of normalization. We do so by employing data augmentation techniques.

Data augmentation is a method used to increase the amount of training data for NLP systems by adding slightly modified copies of already existing data or newly created synthetic data.
In particular, it has been shown to improve performance when a limited amount of labeled samples are available \citep[see][for overviews]{feng-etal-2021-survey,li2022data}. 
Different data augmentation methods exist; most relevant to the current work are noising-based methods such as inserting and changing words. These methods not only expand the amount of training data but also improve model robustness.

\citet{bergmanis2020robust} used an augmentation approach to improve machine translation systems’ performance when faced with orthographic variations (such as unintentional misspellings and deliberate spelling alternations) of Latvian, Estonian and Lithuanian words.
Their results show that, when tested on noisy data, systems trained using adversarial examples performed almost as well as when translating clean data, achieving an improvement of 2-3 BLEU points over the baseline. 
The current study follows a similar approach, but applies it to two NLP tasks for Nigerian Pidgin.

\section{Nigerian Pidgin orthography}

Nigerian Pidgin (commonly referred to simply as `Pidgin' and otherwise known as `Naija') is an English-based contact language that developed as a result of European contact with West African languages. Pidgin incorporates syntax and vocabulary primarily from English, Portuguese and Nigeria’s indigenous languages, as well as new vocabulary that is unique to Nigerian Pidgin.

Like many other pidgin and creole languages, Nigerian Pidgin is a predominantly spoken language. Attempts to write the language go back to the late 18th century \citep{ofulue2010}; nevertheless, it is considered to be fairly young as a written language, since it is still in the process of orthographic standardization and normalization. 
In 2009, the Naijá Langwej Akedemi (NLA) proposed a harmonized orthography
; more recently, \citet{mensah2021towards} published a new proposal for a working orthography of Nigerian Pidgin.
Nevertheless, these orthographies have not yet been adopted by non-linguist Pidgin speakers. Rather, they have developed their own ways of representing their unstandardized language. A gap has thus developed between experts and users, and this may widen in the coming years \citep{deuber2007dynamics}, given that orthographic standardization on the basis of linguists' proposals is not in sight, and that Nigerian Pidgin is not being taught in school. Instead, Nigerian Pidgin speakers are taught in English in school, which likely also influences how these speakers write in Pidgin.

Nigerian Pidgin typically uses a Latin-based alphabet similar to English. Crucially, orthographies tend to be phonetically based with varying degrees of anglicized spellings \citep[see][for related discussions of various Pidgin orthography proposals]{esizimetor2009what,ojarikre2013perspectives}; that is, words are typically spelled and written as pronounced according to the sound patterns of Nigerian Pidgin. However, as we will see in Section \ref{sec:qualanalysis}, variations can still occur in phonetic-based orthographies, resulting in both inter-textual (between texts written by different authors) and intra-textual (within texts written by single authors) orthographic variation. Such inconsistencies in the orthography increase data sparseness and noise \citep{lewis2010haitian}, which affect language models. 

In the approach outlined in this paper, we generate ``adversarial'' training data to be able to train the model to deal with orthographic variation. This effort requires a deeper understanding of the types of variations that occur. Section \ref{sec:qualanalysis} presents a qualitative analysis of common orthographic variations in Nigerian Pidgin.
The remaining subsections then present our approach to create synthetic data with more orthographic variation.




\textbf{}
\begin{table}[t]
\centering
\resizebox{\columnwidth}{!}{
\begin{tabular}{llll}
\toprule
\textbf{Type}          & \textbf{Subtype}       & \textbf{Position} & \textbf{Example}     \\
\midrule
Alternation  & c / k         & \text{initial} & \textbf{c}arry - \textbf{k}arry      \\
              & a / o        & \text{medial} & c\textbf{a}ll - c\textbf{o}ll        \\
              & y / i         & \text{final} & b\textbf{y} - b\textbf{i}            \\
              & e / i        & \text{medial} & d\textbf{e}stroy - d\textbf{i}stroy  \\ 
\midrule
Conversion    & au / o        & \text{medial} & bec\textbf{au}se - bik\textbf{o}s    \\
              & ee / i        & \text{all} & s\textbf{ee} - s\textbf{i}           \\
              & ea / i        & \text{medial} & r\textbf{ea}ch - r\textbf{i}sh       \\
              & eo / i        & \text{medial} & p\textbf{eo}ple - p\textbf{i}pol     \\
              & th / d        & \text{initial} & \textbf{th}e - \textbf{d}i           \\
              & th / t        & \text{all} & \textbf{th}ing - \textbf{t}in        \\  
              & ng / n        & \text{final} & thi\textbf{ng} - ti\textbf{n}        \\  
              & ph / f        & \text{medial} & pro\textbf{ph}et - pro\textbf{f}et    \\
              & wh / w        & \text{initial} & \textbf{wh}en - \textbf{w}en          \\
              & ch / sh       & \text{final} & tea\textbf{ch} - ti\textbf{sh}           \\   
\midrule
Transcription & ble / bol     & \text{final} & trou\textbf{ble} - tro\textbf{bol}   \\
              & ple / pol     & \text{final}   & peo\textbf{ple} - pi\textbf{pol}    \\  
              & er / a        & \text{final} & wheth\textbf{er} - wed\textbf{a}    \\         
              & ight / ite    & \text{final}     & n\textbf{ight} - n\textbf{ite}    \\               
\midrule
Deletion      & e / -  & \text{medial} & diff\textbf{e}rent - difren \\ 
              & e / -   & \text{final} & com\textbf{e} - kom         \\
\bottomrule
\end{tabular}}
\caption{Types of orthographic variation in Nigerian Pidgin.
\textit{Subtype} refers to specific instances of the variation change. 
}\label{tab:variationclass}

\end{table}

\begin{figure*}[ht]
\centering
\includegraphics[width=1.05\textwidth]{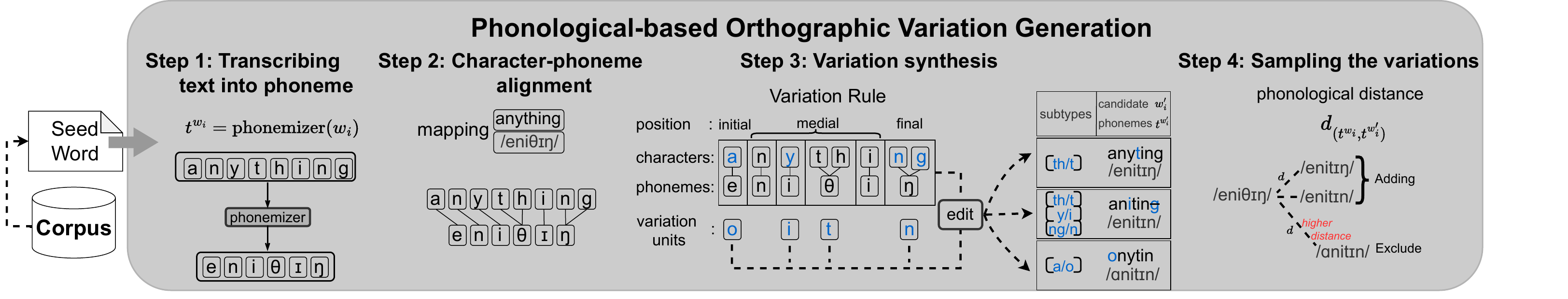}
\caption{ \small Diagram depicting the process of orthographic variation generation for enriching the text corpus.
The seed words are transcribed into phoneme sequences (Step 1), and character-phoneme pairs are aligned (Step 2). 
Next, variants are generated based on the rules (Step 3).
We then measure the phonological distance between the word and the heuristic-generated variation candidates upon their phonemes, denoted as $d(t^{w_{i}},t^{w'_{i}})$ (Step 4).
}
\label{fig:pipeline}
\end{figure*}

\subsection{Types of orthographic variation in Nigerian Pidgin}\label{sec:qualanalysis}

\begin{table}[t]
\centering
\resizebox{1\columnwidth}{!}{
\begin{tabular}{lrrrr} \hline
             & \textit{because} & \textit{bikos} & \textit{cause} & \textit{cos} \\ \hline
Bible        & 265     & 1,108 & 0   & 0     \\
JW300 & 1,504     & 1,613     & 23   & 0    \\ 
Naija Treebank & 243     & 0     & 7   & 3    \\ 
\hline
\end{tabular}
}
\caption{Orthographic variation of the word \textit{because}, along with raw counts per dataset.}\label{tab:variationbecause}
\end{table}

We analysed texts from three parallel datasets: the Bible, JW300 \citep{agic2019} and the Naija Treebank \citep{caron2019surface}. Table \ref{tab:variationclass} presents the types of orthographic variation patterns that were found in this dataset.\footnote{\pinjie{Throughout the development of our framework, the generated variations and examples incorporating Pidgin were evaluated by native Pidgin speakers.}} Note that this list is not exhaustive but rather represents the most common or remarkable variations encountered in the dataset.
\pinjie{See also Appendix~\ref{appendix_variation_types} for details on the construction of the variation types.}

We identify four main classes of systematic variations that occur in the data: (i) alternation between similar sounds; (ii) conversion of digraphs into a single letter or alternate digraphs; (iii) phonetic transcription of (blended) letter pairings; and (iv) deletion of silent letters. Table \ref{tab:variationclass} presents examples of each of these classes. 
The variations often occur at a specific position within a word. 
For instance, the syllable
\textit{ble} can be transcribed as \textit{bol} only when it occurs at the end of a word, as in \textit{trouble}.

Oftentimes a word is characterized by more than one orthographic change occurring. For example, \textit{because}-\textit{bikos} is characterized by alternation (e/i; c/k), conversion (au/o), and deletion (e/-).


Crucially, it should be noted that the variations all have phonetic origins. For example, 
the alternation between /c/ (as in ``carry'') and /k/ (as in ``kid'') can be attributed to both sounding as voiceless velar plosives, and the conversion of the front vowel spelling /ee/ to /i/ can be attributed to these having similar sounds in the Pidgin pronunciation of certain words.

The data is characterized by both intra-textual variation (i.e.~within texts written by single authors) and inter-textual variation (i.e.~between different sources). 
This is illustrated in Table \ref{tab:variationbecause}: in the Bible, the connective `because' is spelled as either \textit{because} or \textit{bikos} (intra-textual variation), whereas in the Naija Treebank, we encounter new variants of the same word (inter-textual variation). 
It is important to note that these variations are not strict rules; different writers might adhere to different rules and thus one word can be written in different ways. For example, the word ``thing'' might be written by one writer as \textit{tin} and by another as \textit{ting} (thus not applying the `ng/n` conversion). Due to a lack of standardized orthography,  there is often not one right variant of a word; rather, we should think of variation candidates as being more, or less, plausible.

Based on the patterns that the qualitative analysis revealed, we develop a framework that exploits the phonological properties of the Pidgin and English words to create orthographic variations. This process is explained in Section \ref{sec:variationcreation}. Section \ref{sec:criteria} then presents how these generated variations were used to augment training data.


\subsection{Synthesizing variation via phonological distance}\label{sec:variationcreation}

The key to synthesizing variants of the words found in the Pidgin dataset is to consider the phonetics of these words. This is because the orthographic variations of Pidgin words tend to originate from those words being written as they are pronounced according to the sound patterns of Nigerian Pidgin.
Based on this insight, we designed an approach that considers the phonemes of the words to generate variants that sound near-identical, but are spelled differently. 

Generating orthographic variations could be done in various ways. One option is to convert phoneme sequences into spelling variants generatively. 
However, phoneme-to-grapheme models usually also rely on already having a model of acceptable word spellings, which is not available in our setting.
We therefore take a different approach: 
we observe what variation exists at the level of orthography, then generate rules based on the orthographic forms, and check the generated variants via pronunciation distance (relying on a grapheme-to-phoneme conversion tool for English). 
These variation rules were based on the qualitative analysis of the dataset described in Section \ref{sec:qualanalysis}. 

In what follows we describe how the framework generates variations. Figure \ref{fig:pipeline} depicts the pipeline for this process. \pinjie{Note that, although this framework was designed for Nigerian Pidgin, it likely can be adapted for other Pidgin-based languages as well, as long as these languages also exploit the phonetic writing system of their lexifier.}



\paragraph{Step 1: Transcribing text into phonemes}



Most orthographies stem from changes at the phonetic level. 
Correctly identifying the character corresponding to the phonetic sound is a crucial step toward accurate word synthesis, as direct substitution based on subwords itself, like $au / o$ often results in variations that do not exist. 
To stimulate the process of creating variation, our framework edits the lexicon by considering acoustic features. 
We first access the sound of a word by using the phonemization tool `phonemizer` \cite{Bernard2021}, which transcribes written words into a series of phonemes that are consistent with English pronunciation rules -- we found that this works well also for Pidgin words.
Given a word $w_{i}$, appearing at \emph{i}-th of a sentence, we obtain the transcription for the Pidgin word, denoted as $t^{w_{i}}$.
In Figure \ref{fig:pipeline}, this is illustrated with the word 'anything', which is transcribed as /\textipa{eniTIN}'/.


\paragraph{Step 2: Character-phoneme alignment}

We adopt GIZA++ \cite{och-ney-2003-systematic} to align the acoustic symbols and the corresponding characters of each word in the given corpus.\footnote{
We trained the \hyperlink{https://github.com/sillsdev/giza-py}{aligner} for 10 iterations using the IBM4 model.  
We utilized the character and phoneme as the training examples.} Due to errors related to missed one-to-many and many-to-one alignments, we first conducted a preprocessing step to merge certain symbols that belong together as a unit, such as /\textipa{o:}/ on the phonetic side, or `th', `ng' on the spelling side, as these would have to be aligned to single symbols such as /\textipa{T}/ or /\textipa{t}/ and /\textipa{N}/ or /\textipa{n}/, respectively, see also Figure \ref{fig:pipeline} for an example. 
The aligned pairs of word and phoneme sequence serve as input for Step 3.





\paragraph{Step 3: Variation synthesis}

To generate orthographic variation candidates of words found in our Pidgin dataset, we created a set of variation rules. These rules were based on the qualitative analysis of the data, as shown in Table \ref{tab:variationclass}. 
When multiple rules could apply to a single word (e.g., in the case of `anything' four rules could apply; see Table \ref{tab:variationclass}), we synthesized variations with different combinations of the rules, see Table \ref{tab:example_similarity}.

We note that the rules overgenerate in some cases; that is, they might in some cases strongly change the pronunciation of a word, e.g., `anything' would result in the alternative spelling `onytin' after the application of several rewriting rules. This variant is however very implausible to occur in Pidgin writing, as the resulting pronunciation in this case would deviate quite a lot from the original pronunciation. This is addressed in Step 4. 


\begin{table}[t]
\centering
\begin{tabular}{l | l c c}
\toprule
{\bfseries Word} & {\bfseries Variation} & {\bfseries LD} & {\bfseries PWLD} \\ \midrule
\multirow{3}{*}{\textsc{Because   }} & bikos & 5 & 0.65 \\
 & cause & 2 & 1.32 \\
 & cos & 5 & 1.36 \\
\midrule

\multirow{3}{*}{\textsc{Anything  }} & anyting & 1 & 0.15  \\
 & anitin & 3 & 0.36 \\
  & onytin & 3 & 0.44 \\

\bottomrule
\end{tabular}
\caption{{Example of pairwise similarity between word and the resulting variations.} 
LD=Levenshtein Distance; PWLD=Phonologically Weighted LD. 
}\label{tab:example_similarity}
\end{table}

\begin{table*}[th]
\small
\centering
\resizebox{\textwidth}{!}{
\begin{tabular}{m{6cm}m{6cm}m{6cm}} 
\toprule
\textbf{Original sample} & \textbf{Variation-enhanced} & \textbf{English translation} \\
\midrule
(\emph{1}) E \textbf{\hlc[babypink]{\textit{come}}} \textbf{\hlc[babypink]{\textit{later}}} dey serve as \textbf{\hlc[babypink]{\textit{pioneer}}} .& 
E \textbf{\hlc[babyblue]{\textit{kome}}} \textbf{\hlc[babyblue]{\textit{lata}}} dey serve as \textbf{\hlc[babyblue]{\textit{pionir}}} .
& \emph{Later, he began serving as a pioneer .}\\
\midrule
(\emph{2}) We \textbf{\hlc[babypink]{\textit{come}}} \textbf{\hlc[babypink]{\textit{later}}} learn for our new place sey if we \textbf{\hlc[babypink]{\textit{want}}} \textbf{\hlc[babypink]{\textit{preach}}}, e \textbf{\hlc[babypink]{\textit{better}}} to go area wey get \textbf{\hlc[babypink]{\textit{another}}} priest .
 & 
We \textbf{\hlc[babyblue]{\textit{kom}}} \textbf{\hlc[babyblue]{\textit{lata}}} learn for our new place sey if we \textbf{\hlc[babyblue]{\textit{wont}}} \textbf{\hlc[babyblue]{\textit{prich}}}, e \textbf{\hlc[babyblue]{\textit{betta}}} to go area wey get \textbf{\hlc[babyblue]{\textit{anotha}}} priest . &  \emph{We later learned at our new place, that if we want to preach, it is better to go to an area with another priest .}
\\
\midrule
(\emph{3}) Wetin we fit do so \textbf{\hlc[babypink]{\textit{that}}} our \textbf{\hlc[babypink]{\textit{character}}} go dey  \textbf{\hlc[babypink]{\textit{better}}} ?  & 
Wetin we fit do so \textbf{\hlc[babyblue]{\textit{dat}}} our \textbf{\hlc[babyblue]{\textit{karakter}}} go dey \textbf{\hlc[babyblue]{\textit{betta}}} ?&  \emph{What can we do to improve our character ?}
\\

\bottomrule
\end{tabular}}
\caption{\small {Examples of orthographic variations generated by our framework, based on sentences from JW300.} }\label{tab:sentence_example}
\end{table*}

\paragraph{Step 4: Sampling the variations}

As a result of Steps 1-3, we now have a set of generated variation candidates for each seed word, including some variants that may be implausible as their pronunciation might not fit well with the original words' pronunciation. 
To filter out such poor orthographic variation candidates, we automatically transcribe each generated variant to its phonetic transcription using again the `phonemizer' tool \cite{Bernard2021}. We then measure the distance between the seed word's pronunciation and the candidate's pronunciation using Phonologically Weighted Levenshtein Distance (PWLD) \cite{abdullah21_interspeech}. The PWLD metric assigns a lower distance to similar phonemes and a larger distance to  dissimilar ones by introducing a weight term. The weight term defines how similar phonetic sounds $a$ and $b$ are in the acoustic space. 
%
The original weight distance matrix was initially learned from English; we edited it slightly to fit permissible pronunciation variations in Pidgin for the generated candidates.
Specifically, we annotated 400 synthesized variants as `good' or `bad' variations, and adjusted the distance feature based on these labels. 

The PWLD metric allows us to rank the generated variation candidate $w_{i}'$ according to their similarity in pronunciation to the original word. 
\begin{equation}\label{eq:d}
d_{(w_{i},w_{i}')}=\text{PWLD}(t^{w_i}, t^{w_i'})
\end{equation}

Note that the distance is measured at the level of the word's transcription ($t^{w_{i}'}$) and the variation candidate's transcription ($t^{w_{i}'}$).
We sample the variations by the normalized probability using \emph{inverse-distance weighting} \cite{lu2008adaptive}. 
The probability to certain variation $w_{i}'$ is normalized by the summation of the inverse distances respective to all variation candidates, defined as\footnote{we write the distance as $d_{w_{i}'}$ for simplicity. It is equivalent to $d_{(w_{i}, w_{i}'})$. This also holds for $d_{w_{j}'}$ in Formula 3.}: 

\begin{equation}\label{eq:p}
p_{(w_{i},w_{i}')} = \frac{d_{w_{i}'}^{-1}}{Z}
\end{equation}

where $Z$ denotes the summation of inverse PLWD scores with regards to all candidates:

\begin{equation}\label{eq:z}
Z = \Sigma_{j} \frac{1}{d_{w'_{j}}}
\end{equation}

The normalization over distance ensures that a higher probability is assigned to candidates with lower distances. This allows frequently selected variants that are pronounced similarly to the original word, while preventing dissimilar ones from being selected.

\subsection{Orthographic variation augmentation}
\label{sec:criteria}
We introduce an \emph{orthographic variation augmentation} approach, which adds sentences with synthesized spellings to the training data.
Consider a corpus with a total of $m$ samples, denoted as $\{x_i\}_{1}^{m} \in D$. We generate the corresponding variation-enhanced versions $x_i'$ by randomly selecting $K$ sentences, $\{x_{1}, ..., x_{k}\}$ and greedily generating a spelling variant for all words in that sentence. This results in $k$ variation-augmented sentences forming the augmented data $D' = \{x_{1}', ..., x_{k}'\}$. 
This augmented data complements the training data, creating $D \cup D'$. Table \ref{tab:sentence_example} presents an example of our approach to generating variations in the context of sentences. 
It is worth noting that a single word can have different generated variants in $D'$ as these variants are sampled from the distance-normalized probability distribution.

Our proposed framework is not restricted to a particular task but is applicable to any NLP task that involves Pidgin data. 
In the next two sections, we explore the utilization of augmented data in two critical NLP tasks: sentiment analysis and machine translation.

\section{Sentiment analysis experiment}

\subsection{Dataset, network and training details}
Table \ref{tab:data} describes the datasets used in our experiments. 
For the task of sentiment analysis, we use the NaijaSenti dataset \citep{muhammad-etal-2022-naijasenti}, which comprises a three-way data split (6.7K/0.6K/1.2K).

We use RoBERTa \cite{DBLP:journals/corr/abs-1907-11692} and BERT \cite{devlin-etal-2019-bert} in base versions. We employed AdamW optimization \cite{loshchilov2018decoupled} with a learning rate of 0.0001, and we found that a smaller learning rate helped prevent overfitting in the downstream task.

We compare three models.
The first model, \textit{PgOnly}, is trained on NaijaSenti for 5 epochs with a mini-batch of 32. 
The second model is a fine-tuned model (\textit{FT}), which is trained on English and fine-tuned on Nigerian Pidgin. 
Finally, the third model is an orthographically augmented model (\textit{Orth-Augm}) which, like FT, is trained on English and fine-tuned on Nigerian Pidgin, but the fine-tuning now also includes orthographic variations, which are appended to the training data.
Our orthographic variation augmentation approach samples K=100 variants. We re-ran the model 6 times with different random seeds and present averaged results.
All models are evaluated using the F1 score.

\begin{table}[t]
\small
\centering \tabcolsep=3pt
\resizebox{\columnwidth}{!}{
\begin{tabular}{ll  c  r l}
\toprule \textbf{Task}&
{\bfseries Corpus} & {\bfseries Language} & {\bfseries $|$Train$|$} & {\bfseries Domain} \\
\midrule
SA & NaijaSenti  & \textsc{Pg.} & $8,524$ & social media \\
\midrule
&Bible & \textsc{Pg.}, \textsc{En.} & $29,737$ & religious \\
MT&JW300  & \textsc{Pg.}, \textsc{En.} & $20,218$ & religious \\
&\text{Naija Treebank}   & \textsc{Pg.}, \textsc{En.} & $9,240$ & misc. \\
\bottomrule
\end{tabular}}
\caption{{Overview of Pidgin datasets used in the current work.} \textsc{Pg.}: data in Nigerian Pidgin; \textsc{En.}: translations in English language. |Train| = Data size in number of sentences.
}\label{tab:data}
\end{table}

\begin{table}[t]
\centering
\begin{tabular}{l|ccc}
\toprule
\textbf{Model Type} & \textbf{PgOnly} & \textbf{FT} & \textbf{\textbf{Orth-Augm}} \\
\midrule
\text{BERT} & 71.8 & 79.7 & $\textbf{80.21}_{\pm \text{0.21}}$ \\
\text{RoBERTa} & 68.4 & 80.1 & $\textbf{82.23}_{\pm \text{0.17}}$ \\
\bottomrule
\end{tabular}
\caption{{Results (F1) of sentiment classification on \box0{NaijaSenti} (8.8K; six runs).}
}
\label{tab:classification-results}
\end{table}



\subsection{Main results}\label{sec:results}

Table \ref{tab:classification-results} presents the results on the sentiment analysis task.
Both BERT and RoBERTa models, when trained with orthographic variation augmented data, demonstrated improvements in F1 scores, gaining $+0.51$ and $+2.19$ points over the fine-tuned model (FT), respectively.
We notice the higher improvement between FT and our approach when using RoBERTa in comparison to BERT.
Even though RoBERTa leverages more pre-training data over BERT, the model still shows more improvements when trained with the augmented variations.

The model variant PgOnly is only trained on NaijaSenti and has not seen any English data. PgOnly shows significant gaps compared to models with fine-tuned and variation-augmented training.
This result emphasizes the importance of fine-tuning language models and how such models can be further improved with our generated variations.


\subsection{Ablation study}
This section provides further results on the augmentation effectiveness and shows the advantage of our augmentation approach.

\begin{figure}[t]
\centering
\includegraphics[width=1\columnwidth,height=1\textheight,keepaspectratio]{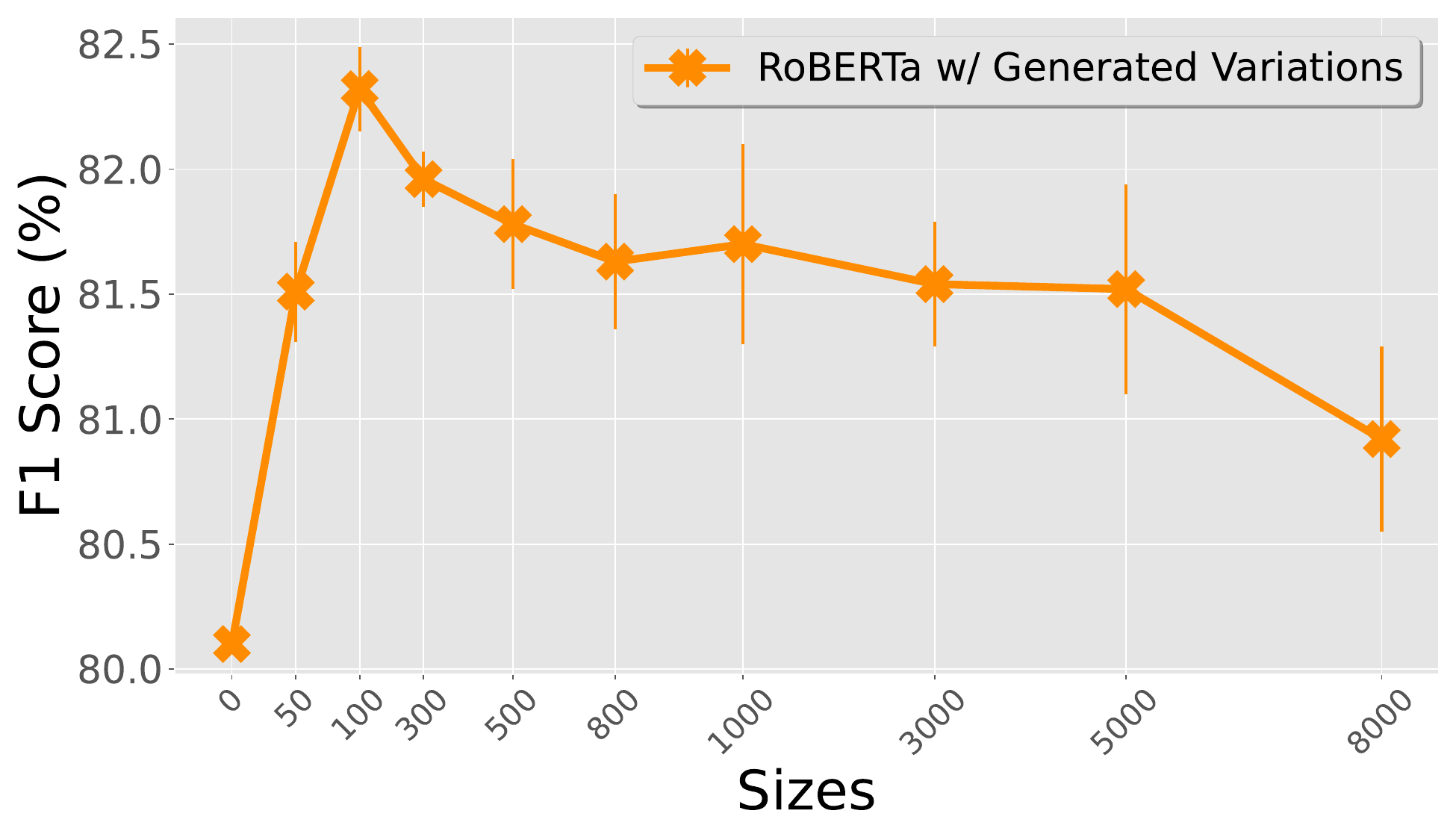}
\caption{Performance on various $K$ augmented sample size. Error bars reflect the standard error over six runs.}
\label{fig:effect_k}
\end{figure}

\paragraph{Effect of $K$ augmented samples.}
To determine the optimal size of augmented samples  $K$ for the sentiment analysis task, we adopt RoBERTa and test various sizes of extra data points within a range from $K$=50 to $K$=8,000.
Figure \ref{fig:effect_k}  presents the results in comparison to the fine-tuned baseline ($K$ size=0).

\begin{figure}[t]
\centering
\includegraphics[width=1\columnwidth,height=1\textheight,keepaspectratio]{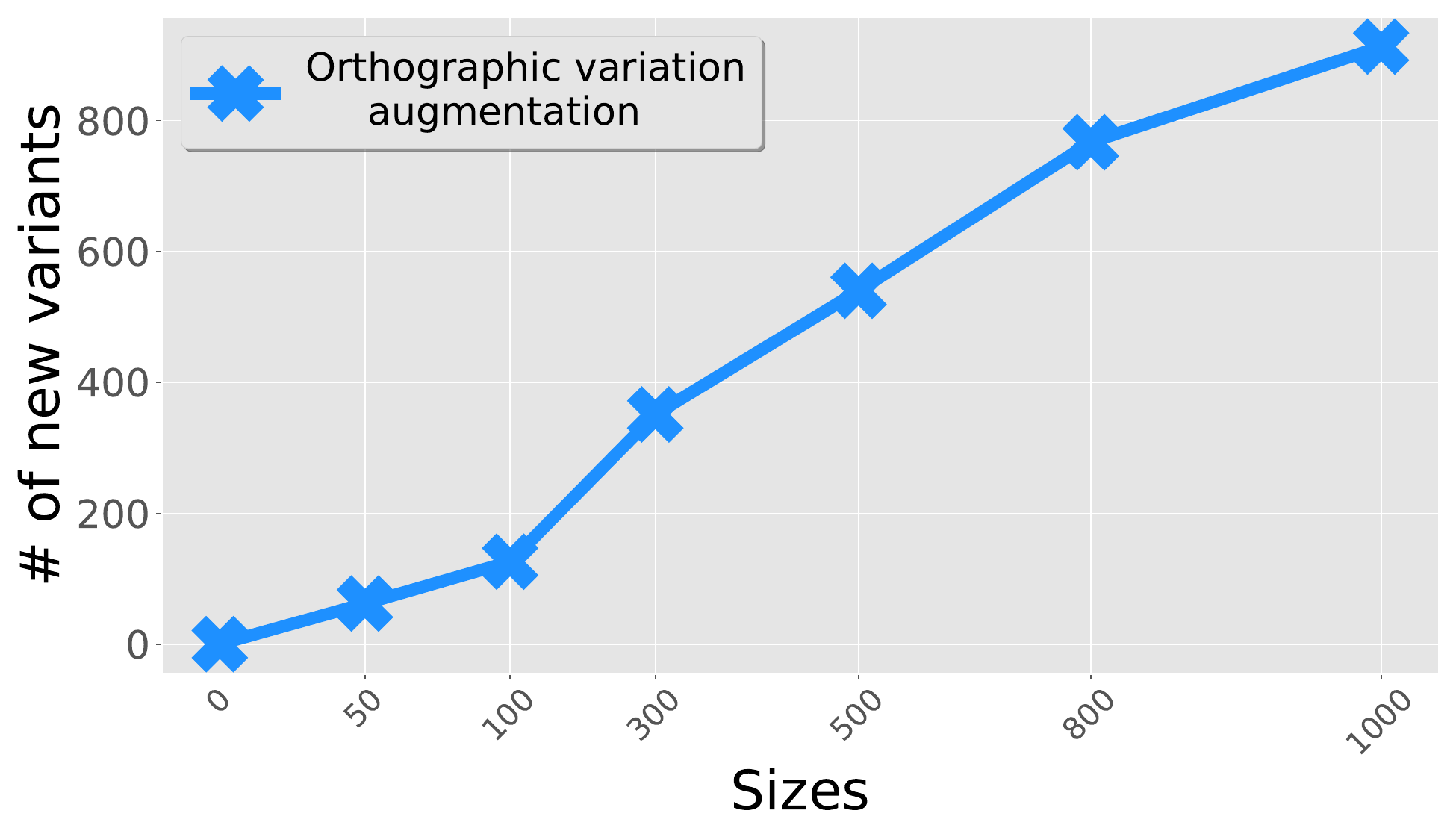}
\caption{Number of new variants from variation-enhanced augmentation data.
}
\label{fig:new_word}
\end{figure}

Figure \ref{fig:effect_k} shows that appending 100 augmented samples leads to a substantial performance boost of $+2.19$ points over the baseline.
However, as more augmented samples are introduced, we see a gradual decline in performance to $81.09$.
We attribute the observed fluctuations to the explanation that, while additional generated variations enrich the diversity of the training data, a relatively small test set does not present such high variability of the real variation distribution, and thus the additional augmentation introduces disproportionally much noise.
Figure \ref{fig:new_word} shows the potential reason for the fluctuation: an increasing number of new variants is introduced as more augmented data is appended to the training dataset.
Nevertheless, even with 8K samples, the improvement in performance compared to the baseline is close to $1$ point in F1 score (as shown in Figure \ref{fig:effect_k}). 
In sum, we find that controlling the size of the synthesized data influences the level of generalization achieved, but regardless of the size $K$, adding orthographic variants to the dataset improved the baseline model performance on NaijaSenti.

\paragraph{Cross-entropy of variation-augmented data.}
We further explore training stability by examining the cross-entropy of the model's prediction $p(y|x)$, in the context of sentiment analysis.
Table \ref{tab:entropy} demonstrates the cross-entropy in the training epochs, as a crucial metric under cross-validation to gauge the effectiveness of our classification model.
Cross-entropy informs us on how well the model's predictions match the ground-truth class labels. 
Measuring cross-entropy is done as an alternative to accuracy, which is less applicable to smaller datasets. 
The results in Table \ref{tab:entropy} show that models trained using samples with orthographic variations are characterised by lower cross-entropy compared to the counterpart models that are not trained with appended orthographic variation samples.
This indicates that our orthographic variation augmentation approach leads to more accurate sentiment analysis results.



\begin{table}[t]
\centering
\resizebox{\columnwidth}{!}{
\begin{tabular}{l|ccccc}
\toprule
& \textbf{EP. 1} & \textbf{EP. 3} & \textbf{EP. 5} & \textbf{\textsc{EP. 10}} & \textbf{\textsc{EP. 20}} \\

\midrule
\text{no augm.} & .98 & .99 & 1.04 & .97 & .94 \\
\text{Ortho. augm.} & .69 & .58 & .55 & .53  & .49 \\
\bottomrule
\end{tabular}}
\caption{Cross-entropy results from trained models for various training epochs. 
}
\label{tab:entropy}
\end{table}
\section{Machine translation experiment}

\subsection{Dataset, networks and training details}
We leverage \textsc{T5 Base} \cite{JMLR:v21:20-074} for the JW300 translation benchmark, which consists of 29K training samples. We perform bi-directional translation for the parallel English-Pidgin datasets, which are the Bible, JW300 \citep{agic2019} and the Naija Treebank \citep{caron2019surface}. 
All model variants are evaluated on the test set using BLEU scores.

We fine-tuned the T5-base model with additional real data samples (DataAug) for 20 epochs with a batch size of 64.
We configured the sequence length to be 196 characters and used the AdamW optimizer \cite{loshchilov2018decoupled} with a learning rate of 0.0001. 

We appended the orthographic variation samples as augmentation, using $K$=20,000 samples. Depending on the model type (only JW300 or a combination of JW300 and the Bible or Treebank), the augmentation samples were drawn from either a single source or from multiple sources. 


\begin{table}[t]
\centering
\resizebox{\columnwidth}{!}{
\begin{tabular}{lll}
\toprule
\textbf{Model Type} & \textbf{EN.-PG.} & \textbf{PG.-EN.}\\ 
\midrule
\emph{\textsc{Baseline (JW300)}} & \text{34.87} & \text{32.56} \\
\midrule
\emph{\textsc{DataAug}} \\
\textsc{\hspace{3mm}JW300+Bible} & \text{35.13} & \text{33.79} \\
\textsc{\hspace{3mm}JW300+Bible+TreeBank} & \text{35.34} & \text{33.41} \\ 
\midrule
\emph{\textsc{with orthographic variation}} \\
\textsc{\hspace{3mm}JW300} & $\text{35.14}_{\pm \text{0.13}}$ & $\text{33.52}_{\pm \text{0.18}}$ \\
\textsc{\hspace{3mm}JW300+Bible} & $\text{35.55}_{\pm \text{0.14}}$ & $\text{33.91}_{\pm \text{0.22}}$ \\
\textsc{\hspace{3mm}JW300+Bible+TreeBank} & $\textbf{35.61}_{\pm \text{0.23}}$ & $\textbf{33.95}_{\pm \text{0.19}}$ \\
\bottomrule
\end{tabular}}
\caption{Results (BLEU scores) of traditional data augmentation (DataAug) and the proposed argumentation with orthographic variation, tested on JW300 test set.
}
\label{tab:translation-results}
\end{table}



\subsection{Main results}
Table \ref{tab:translation-results} shows that our orthographic variation augmentation approach leads to improvements in BLEU scores in both translation directions.
The results show that this also surpasses the standard data augmentation technique in performance.
For DataAug (including real samples of additional datasets), the model's performance benefits from injecting more accurately labeled real training samples, leading to an improvement in BLEU points. 
We observe further improvement by using our orthographic variant generation approach, suggesting that the augmented samples are an effective augmentation approach to
enrich the dataset.

\subsection{Ablation study}



\paragraph{Cross Domain Generalization on Treebank}

Given that orthographic variation is even more likely to occur between texts than within texts, we test the effect of orthographic variation augmentation on a test set from an unseen corpus. Specifically, we investigate the cross domain transfer of a model trained on the JW300 (with and without orthographic variations) and tested on the Naija Treebank test set.

\begin{figure}[t]
\centering
\begin{subfigure}[b]{0.43\textwidth}
       \includegraphics[width=1\linewidth]{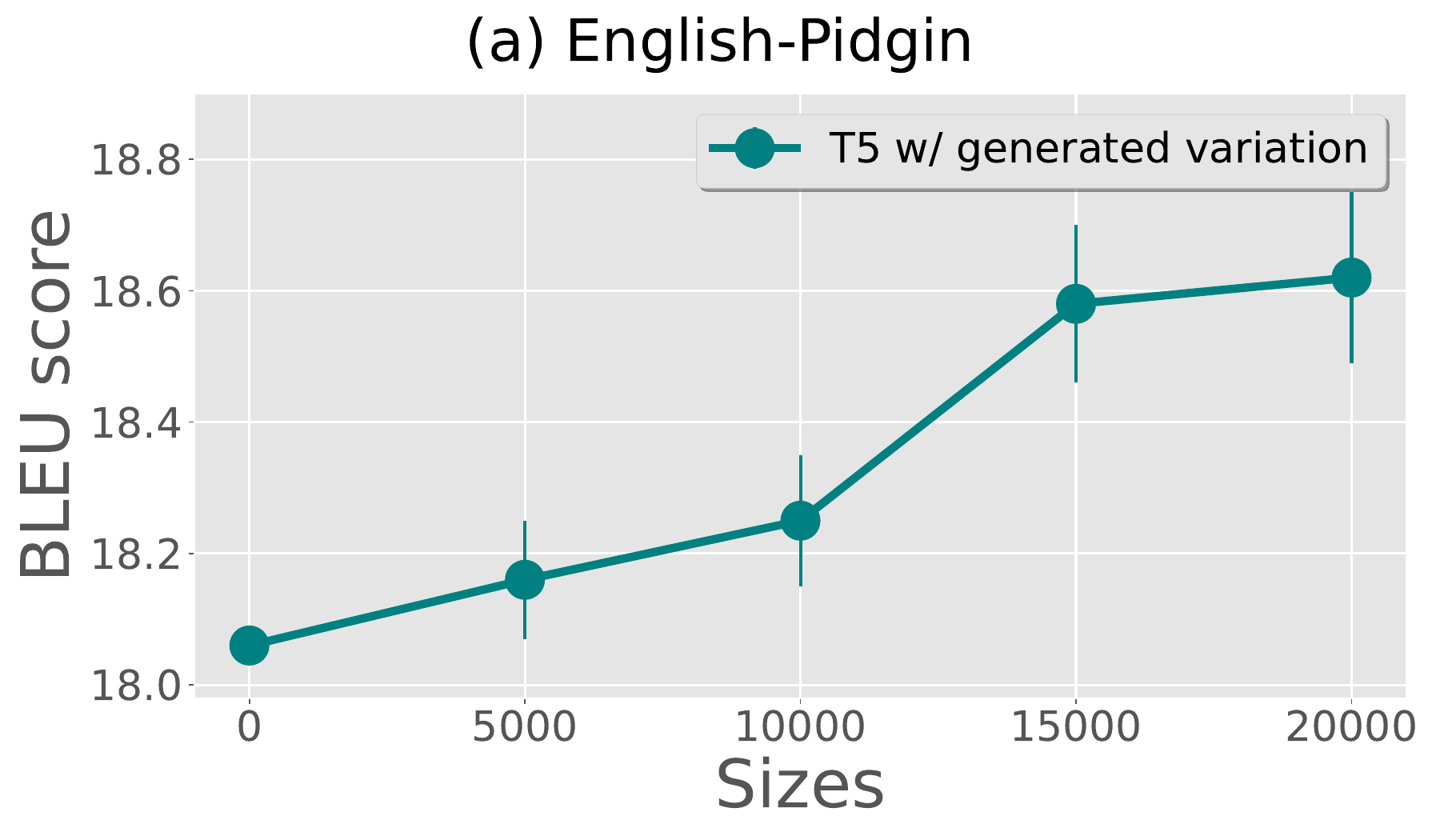}
       \label{fig:en2pcm_bleu}
    \end{subfigure}
    \begin{subfigure}[b]{0.43\textwidth}
       \includegraphics[width=1\linewidth]{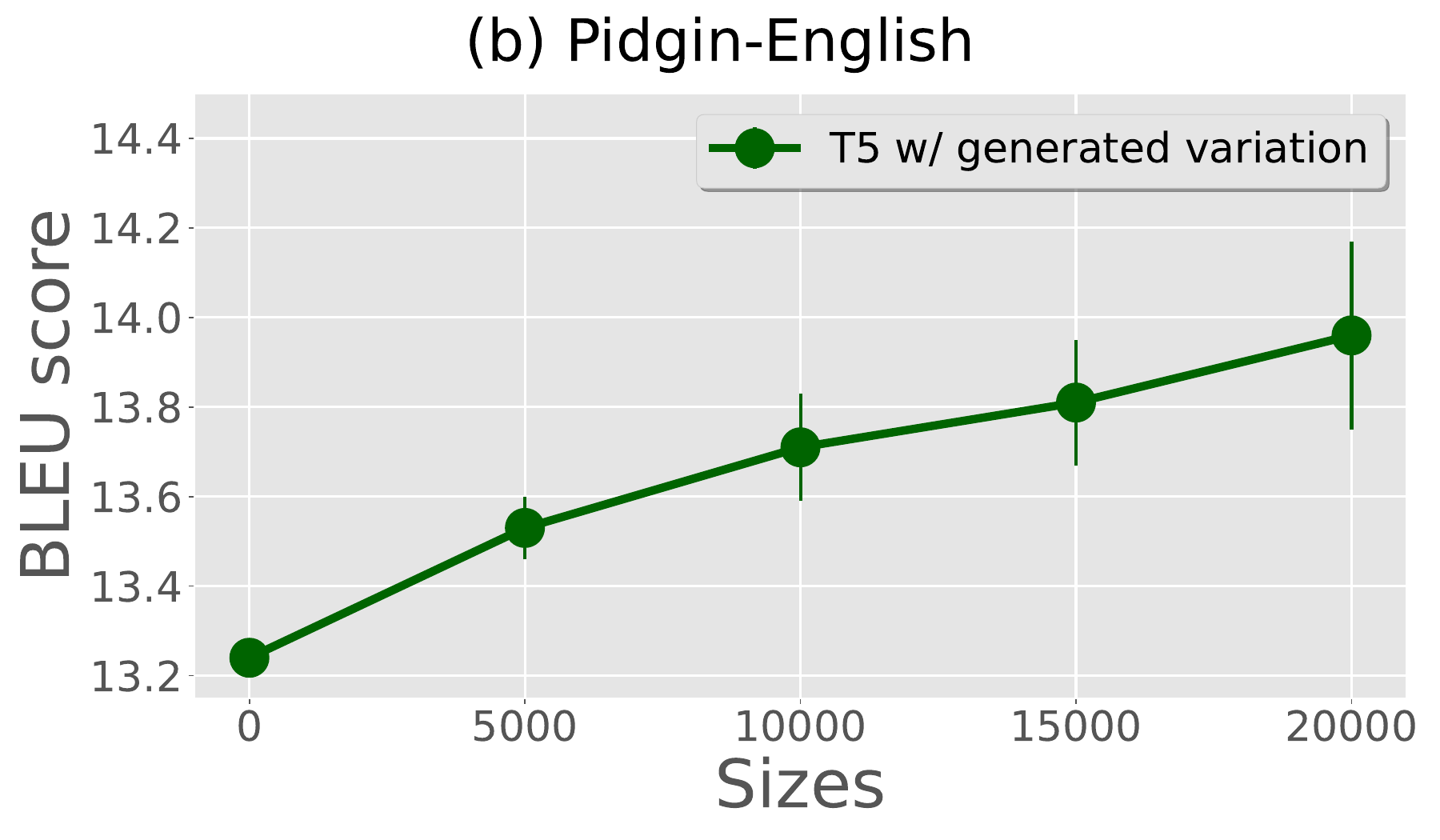}
       \label{fig:pcm2en_bleu}
    \end{subfigure}
\caption{Generalization from JW300 (training) to Naija Treebank (testing).
We show the improvements in BLEU scores while varying the number of added orthographic variants.}
\label{fig:domain_shift_on_jw300}
\end{figure}

Figure \ref{fig:domain_shift_on_jw300} shows that our augmentation approach improves the performance of the machine translation model, on the unseen Naija Treebank during training, leading to an zero-shot generalization along with more augmented samples.
In fact, the results show that the higher the $K$, the higher the model performance over the baseline $K=0$ where no variation-enhanced sentences are used, further supporting the effectiveness of the variation augmentation.

\paragraph{New variants improve generalization}

The data augmentation approach introduces new word variants to the training data; naturally, the number of new variants introduced increases when the size of the augmented data increases.
To understand how these new variants affect the generalization of the model, we correlate the number of new variants for various $K$ sizes with the performance improvement in the domain shifting setting.
Table \ref{tab:new_words} shows that with a higher number of new variants, the results also show a performance improvement on the Naija Treebank's test set.
We attribute this phenomenon to two factors: (1) certain newly created variations are absent in the current test split but are indeed found in real Pidgin corpora, as exemplified by `everytin'; (2) in other cases, such as `piple' and `pipl', these variants feature minor lexical alterations compared to the variant `pipol' that is actually found in the dataset.
Even though certain generated variants, such as `piple' and `pipl', are classified as new variants, parts of these created variants are indeed present in the original text (e.g.~`pip' in the three variants `piple', `pipl', and `pipol') and the model might capture nearly identical semantics due to their lexical overlap. 
This could lead to the model being able to generalize to previously unseen tokens when conducting inference on the test splits.

\begin{table}[t]
\centering
\resizebox{\columnwidth}{!}{
\begin{tabular}{l|cccc}
\toprule
 & \textbf{5K} & \textbf{10K} & \textbf{15K} & \textbf{20K} \\
\midrule
\text{New variants} & \text{561} &  \text{659} &  \text{706} &  \text{745}  \\
\text{Improvement in BLEU - EN-PG} & 0.10 & 0.19  & 0.52  & \textbf{0.56}  \\
\text{Improvement in BLEU - PG-EN} & 0.29 & 0.47  & 0.57  & \textbf{0.72}  \\
\bottomrule
\end{tabular}}
\caption{{
Correlation of new variants and performance improvement on Naija Treebank (testing)}. The augmented data introduces seen as well as previously unseen orthographic variants. Numbers represent the relative improvement in BLEUs on Naija Treebank. 
}
\label{tab:new_words}
\end{table}

\section{Discussion}
\subsection{Overgeneration}
\pinjie{
Although Nigerian Pidgin does not have a standard orthography, there is still a degree of plausibility in the generations that can be found in natural language. Automatically generating orthographic variations can thus lead to implausible variations, which, in turn, can lead to a decrease in generalization. A manual analysis of the data indicated that many of the implausible variations generated by our approach were synthesized when multiple rules were applied on the seed word, and when the pronunciation of a word was strongly affected. For instance, ‘anything’  became ‘onytin’ by applying three different rules. Such candidates are less plausible variations than one that has higher phonological similarity (also see Appendix \ref{appendix_irrelevant_words} on generating irrelevant words). In our framework, we address the issue of overgeneration by sampling the variations through the phonological weighted Levenshtein distance. 
}

\subsection{Generalization to unseen domains}
\pinjie{
The dataset used to identify common orthographic variations consisted of texts from the bible, a religious magazine (JW300), and Nigerian Pidgin conversations (Naija Treebank). The variations observed within these sources appeared to be more author-specific than domain-specific -- that is, the degree and variety of variations often involve specific authors instead of domains. An example of this can be seen in Table \ref{tab:variationbecause}: the Bible and JW300 are from the same domain, but the types of variations and the frequencies are very different. 
\\
The experiments reported in Sections 4 and 5 show that the proposed method can be applied across domains. For sentiment analysis and machine translation tasks, we created separate frameworks on two distinct domains; social media and religion, respectively. Our results showed positive correlations when the increase of the model performance and augmenting the variation-enhance examples (also see Tables \ref{tab:classification-results} and \ref{tab:translation-results}). Our proposed approach thus appears to be domain-agnostic and can be generalized to unseen domains.}

\subsection{Code-switching}

\pinjie{Nigerian Pidgin is an English-lexified language that draws from other local languages as well. Code-switching between Pidgin, English, and local languages is motivated by factors such as formality, setting, interpersonal relations and audience \citep{agbo2020relationship}. 
For example, in a Hausa community, there might be more code-switching with Hausa words, but in a mixed community of Hausa and Yoruba speakers, there might be more code-switching with (Nigerian-)English words. 
Table \ref{tab:sentence_example} presents an example of how English and Pidgin words co-occur in one uttereance: `E come later dey serve as pioneer' translates to English as `Later, he began serving as a pioneer.' In this sentence, the speaker mixes Pidgin words (e.g., `dey') with (Nigerian-)English words (e.g., `serve as pioneer'). This blend of English and Nigerian Pidgin is a form of code-switching commonly observed within the text (although it could be argued that there might not be a Pidgin equivalent of “pioneer” available in the lexicon). 
\\
Our current work proposes a framework to describe various types of orthographic variations commonly found in Nigerian Pidgin texts, and model this orthographic variation. As a result, words like `they' in the dataset might be generated as `dey'. Through this process, the original text is enhanced with more orthographic variations, which  potentially creates the illusion of a higher rate of code-switching than is originally present in the data. \\
However, note that this view on code-switching relies solely on orthographic choices made by the transcriber, whereas code-switching also pertains to vocabulary items that are not dependent on orthographic choices (such as the Pidgin word `wetin' meaning \textit{what}). We refer the reader to \citet{agbo2022language} for more information on code-switching in Nigerian Pidgin.
}

\section{Conclusion}

In this paper, we address the issue of orthographic variation in a mostly spoken language that does not have a standardized orthography in place: Nigerian Pidgin (Naija). 
We provide an analysis of the types of orthographic variations commonly found in Nigerian Pidgin writing. 
Based on this analysis, we propose a novel phonological-based word synthesizing framework to augment the corpus with orthographic variations.
We examine the concept of synthesized variation on two main tasks: sentiment analysis and machine translation. The results demonstrate the effectiveness of adding synthesized orthographic variation to the dataset instead of collecting new samples.

\section{Acknowledgement}
This work was funded by the Deutsche Forschungsgemeinschaft (DFG, German Research Foundation) – Project-ID 232722074 – SFB 1102 Information Density and Linguistic Encoding.

\section{Limitations}
We acknowledge the inherent limitations in our work.
Despite the effectiveness of our orthographic variation generation framework, we still observed the overgeneration of variations that are less likely to occur.
We face a challenge in precisely quantifying the correlation between the model's performance and the extent of overgeneration. The absence of such measurement hinders a comprehensive assessment of the impact of overgeneration on model performance.

Moreover, our evaluation of NaijaSenti is constrained to an in-domain context due to the lack of an out-of-domain corpus that represents stronger or novel variations, potentially resulting in an underestimation of the model's capabilities. 
Additionally, the exploration of alternative data-driven sampling methods is a important option; however, it requires additional non-annotated Pidgin data, which is currently unavailable.

Lastly, our research focuses on Nigerian Pidgin, which is an English-lexified language; this allowed us to exploit the availability of SOTA English-based tools. It is possible that our observations do not generalize well to other languages that are either not English-lexified (due to a less resources being available for other languages), or display more differences compared to their lexifier. 

\section{Bibliographical References}\label{reference}

\bibliographystyle{lrec-coling2024-natbib}
\bibliography{lrec-coling2024-example,languageresource}

\begin{thebibliography}{31}
\expandafter\ifx\csname natexlab\endcsname\relax\def\natexlab#1{#1}\fi

\bibitem[{Abdullah et~al.(2021)Abdullah, Mosbach, Zaitova, Möbius, and
  Klakow}]{abdullah21_interspeech}
Badr~M. Abdullah, Marius Mosbach, Iuliia Zaitova, Bernd Möbius, and Dietrich
  Klakow. 2021.
\newblock \href {https://doi.org/10.21437/Interspeech.2021-678} {{Do Acoustic
  Word Embeddings Capture Phonological Similarity? An Empirical Study}}.
\newblock In \emph{Proc. Interspeech 2021}, pages 4194--4198.

\bibitem[{Adelani et~al.(2021)Adelani, Abbott, Neubig, D{'}souza, Kreutzer,
  Lignos, Palen-Michel, Buzaaba, Rijhwani, Ruder, Mayhew, Azime, Muhammad,
  Emezue, Nakatumba-Nabende, Ogayo, Anuoluwapo, Gitau, Mbaye, Alabi, Yimam,
  Gwadabe, Ezeani, Niyongabo, Mukiibi, Otiende, Orife, David, Ngom, Adewumi,
  Rayson, Adeyemi, Muriuki, Anebi, Chukwuneke, Odu, Wairagala, Oyerinde, Siro,
  Bateesa, Oloyede, Wambui, Akinode, Nabagereka, Katusiime, Awokoya, MBOUP,
  Gebreyohannes, Tilaye, Nwaike, Wolde, Faye, Sibanda, Ahia, Dossou, Ogueji,
  DIOP, Diallo, Akinfaderin, Marengereke, and
  Osei}]{adelani-etal-2021-masakhaner}
David~Ifeoluwa Adelani, Jade Abbott, Graham Neubig, Daniel D{'}souza, Julia
  Kreutzer, Constantine Lignos, Chester Palen-Michel, Happy Buzaaba, Shruti
  Rijhwani, Sebastian Ruder, Stephen Mayhew, Israel~Abebe Azime, Shamsuddeen~H.
  Muhammad, Chris~Chinenye Emezue, Joyce Nakatumba-Nabende, Perez Ogayo, Aremu
  Anuoluwapo, Catherine Gitau, Derguene Mbaye, Jesujoba Alabi, Seid~Muhie
  Yimam, Tajuddeen~Rabiu Gwadabe, Ignatius Ezeani, Rubungo~Andre Niyongabo,
  Jonathan Mukiibi, Verrah Otiende, Iroro Orife, Davis David, Samba Ngom, Tosin
  Adewumi, Paul Rayson, Mofetoluwa Adeyemi, Gerald Muriuki, Emmanuel Anebi,
  Chiamaka Chukwuneke, Nkiruka Odu, Eric~Peter Wairagala, Samuel Oyerinde,
  Clemencia Siro, Tobius~Saul Bateesa, Temilola Oloyede, Yvonne Wambui, Victor
  Akinode, Deborah Nabagereka, Maurice Katusiime, Ayodele Awokoya, Mouhamadane
  MBOUP, Dibora Gebreyohannes, Henok Tilaye, Kelechi Nwaike, Degaga Wolde,
  Abdoulaye Faye, Blessing Sibanda, Orevaoghene Ahia, Bonaventure F.~P. Dossou,
  Kelechi Ogueji, Thierno~Ibrahima DIOP, Abdoulaye Diallo, Adewale Akinfaderin,
  Tendai Marengereke, and Salomey Osei. 2021.
\newblock \href {https://doi.org/10.1162/tacl_a_00416} {{M}asakha{NER}: Named
  entity recognition for {A}frican languages}.
\newblock \emph{Transactions of the Association for Computational Linguistics},
  9:1116--1131.

\bibitem[{Agbo(2022)}]{agbo2022language}
Ogechi~Florence Agbo. 2022.
\newblock \emph{Language Use and Code-switching among Educated English-Nigerian
  Pidgin Bilinguals in Nigeria}.
\newblock Ph.D. thesis, Dissertation, D{\"u}sseldorf,
  Heinrich-Heine-Universit{\"a}t, 2022.

\bibitem[{Agbo and Plag(2020)}]{agbo2020relationship}
Ogechi~Florence Agbo and Ingo Plag. 2020.
\newblock The relationship of nigerian english and nigerian pidgin in nigeria:
  Evidence from copula constructions in ice-nigeria.
\newblock \emph{Journal of Language Contact}, 13(2):351--388.

\bibitem[{Agi{\'c} and Vuli{\'c}(2019)}]{agic2019}
{\v{Z}}eljko Agi{\'c} and Ivan Vuli{\'c}. 2019.
\newblock \href {https://doi.org/10.18653/v1/P19-1310} {{JW}300: A
  wide-coverage parallel corpus for low-resource languages}.
\newblock In \emph{Proceedings of the 57th Annual Meeting of the Association
  for Computational Linguistics}, pages 3204--3210, Florence, Italy.
  Association for Computational Linguistics.

\bibitem[{Ajisafe et~al.(2021)Ajisafe, Adegboro, Oduntan, and
  Arulogun}]{ajisafe2020towards}
Daniel Ajisafe, Oluwabukola Adegboro, Esther Oduntan, and Tayo Arulogun. 2021.
\newblock Towards end-to-end training of automatic speech recognition for
  nigerian pidgin.
\newblock In \emph{ICASSP 2021 - 2021 IEEE International Conference on
  Acoustics, Speech and Signal Processing (ICASSP)}.

\bibitem[{Bergmanis et~al.(2020)Bergmanis, Stafanovi{\v{c}}s, and
  Pinnis}]{bergmanis2020robust}
Toms Bergmanis, Art{\=u}rs Stafanovi{\v{c}}s, and M{\=a}rcis Pinnis. 2020.
\newblock Robust neural machine translation: Modeling orthographic and
  interpunctual variation.
\newblock In \emph{Human Language Technologies--The Baltic Perspective}, pages
  80--86. IOS Press.

\bibitem[{Bernard and Titeux(2021)}]{Bernard2021}
Mathieu Bernard and Hadrien Titeux. 2021.
\newblock \href {https://doi.org/10.21105/joss.03958} {Phonemizer: Text to
  phones transcription for multiple languages in python}.
\newblock \emph{Journal of Open Source Software}, 6(68):3958.

\bibitem[{Caron et~al.(2019)Caron, Courtin, Gerdes, and
  Kahane}]{caron2019surface}
Bernard Caron, Marine Courtin, Kim Gerdes, and Sylvain Kahane. 2019.
\newblock {A surface-syntactic UD treebank for Naija}.
\newblock In \emph{TLT 2019, Treebanks and Linguistic Theories, Syntaxfest}.

\bibitem[{Deuber and Hinrichs(2007)}]{deuber2007dynamics}
Dagmar Deuber and Lars Hinrichs. 2007.
\newblock Dynamics of orthographic standardization in {Jamaican Creole and
  Nigerian Pidgin}.
\newblock \emph{World Englishes}, 26(1):22--47.

\bibitem[{Devlin et~al.(2019)Devlin, Chang, Lee, and
  Toutanova}]{devlin-etal-2019-bert}
Jacob Devlin, Ming-Wei Chang, Kenton Lee, and Kristina Toutanova. 2019.
\newblock \href {https://doi.org/10.18653/v1/N19-1423} {{BERT}: Pre-training of
  deep bidirectional transformers for language understanding}.
\newblock In \emph{Proceedings of the 2019 Conference of the North {A}merican
  Chapter of the Association for Computational Linguistics: Human Language
  Technologies, Volume 1 (Long and Short Papers)}, pages 4171--4186,
  Minneapolis, Minnesota. Association for Computational Linguistics.

\bibitem[{Esizimetor(2009)}]{esizimetor2009what}
David~Oshorenoya Esizimetor. 2009.
\newblock What orthography for naij{\'a}?
\newblock In \emph{Proceedings of the Conference on Nigerian Pidgin, University
  of Ibadan, Nigeria}.

\bibitem[{Eskander et~al.(2013)Eskander, Habash, Rambow, and
  Tomeh}]{eskander2013processing}
Ramy Eskander, Nizar Habash, Owen Rambow, and Nadi Tomeh. 2013.
\newblock Processing spontaneous orthography.
\newblock In \emph{Proceedings of the 2013 Conference of the North American
  chapter of the association for computational linguistics: Human language
  technologies}, pages 585--595.

\bibitem[{Feng et~al.(2021)Feng, Gangal, Wei, Chandar, Vosoughi, Mitamura, and
  Hovy}]{feng-etal-2021-survey}
Steven~Y. Feng, Varun Gangal, Jason Wei, Sarath Chandar, Soroush Vosoughi,
  Teruko Mitamura, and Eduard Hovy. 2021.
\newblock \href {https://doi.org/10.18653/v1/2021.findings-acl.84} {A survey of
  data augmentation approaches for {NLP}}.
\newblock In \emph{Findings of the Association for Computational Linguistics:
  ACL-IJCNLP 2021}, pages 968--988, Online. Association for Computational
  Linguistics.

\bibitem[{Lent et~al.(2021)Lent, Bugliarello, de~Lhoneux, Qiu, and
  S{\o}gaard}]{lent2021language}
Heather Lent, Emanuele Bugliarello, Miryam de~Lhoneux, Chen Qiu, and Anders
  S{\o}gaard. 2021.
\newblock On language models for creoles.
\newblock In \emph{Proceedings of the 25th Conference on Computational Natural
  Language Learning}, pages 58--71.

\bibitem[{Lewis(2010)}]{lewis2010haitian}
William Lewis. 2010.
\newblock {Haitian Creole: How to build and ship an MT Engine from scratch in 4
  days, 17 hours, \& 30 minutes}.
\newblock In \emph{Proceedings of the 14th Annual conference of the European
  Association for Machine Translation}.

\bibitem[{Li et~al.(2022)Li, Hou, and Che}]{li2022data}
Bohan Li, Yutai Hou, and Wanxiang Che. 2022.
\newblock Data augmentation approaches in natural language processing: A
  survey.
\newblock \emph{Ai Open}, 3:71--90.

\bibitem[{Lin et~al.(2023)Lin, Saeed, Chang, and Scholman}]{lin2023low}
Pin-Jie Lin, Muhammed Saeed, Ernie Chang, and Merel Scholman. 2023.
\newblock Low-resource cross-lingual adaptive training for {Nigerian Pidgin}.
\newblock In \emph{Interspeech 2023}, pages 3954--3958.

\bibitem[{Liu et~al.(2019)Liu, Ott, Goyal, Du, Joshi, Chen, Levy, Lewis,
  Zettlemoyer, and Stoyanov}]{DBLP:journals/corr/abs-1907-11692}
Yinhan Liu, Myle Ott, Naman Goyal, Jingfei Du, Mandar Joshi, Danqi Chen, Omer
  Levy, Mike Lewis, Luke Zettlemoyer, and Veselin Stoyanov. 2019.
\newblock \href {http://arxiv.org/abs/1907.11692} {Roberta: {A} robustly
  optimized {BERT} pretraining approach}.
\newblock \emph{CoRR}, abs/1907.11692.

\bibitem[{Loshchilov and Hutter(2019)}]{loshchilov2018decoupled}
Ilya Loshchilov and Frank Hutter. 2019.
\newblock \href {https://openreview.net/forum?id=Bkg6RiCqY7} {Decoupled weight
  decay regularization}.
\newblock In \emph{International Conference on Learning Representations}.

\bibitem[{Lu and Wong(2008)}]{lu2008adaptive}
George~Y Lu and David~W Wong. 2008.
\newblock An adaptive inverse-distance weighting spatial interpolation
  technique.
\newblock \emph{Computers \& geosciences}, 34(9):1044--1055.

\bibitem[{Mensah et~al.(2021)Mensah, Ukaegbu, and Nyong}]{mensah2021towards}
Eyo Mensah, Eunice Ukaegbu, and Benjamin Nyong. 2021.
\newblock Towards a working orthography of {Nigerian Pidgin}.
\newblock \emph{Current Trends in Nigerian Pidgin English}, page 177.

\bibitem[{Muhammad et~al.(2022)Muhammad, Adelani, Ruder, Ahmad, Abdulmumin,
  Bello, Choudhury, Emezue, Abdullahi, Aremu, Jorge, and
  Brazdil}]{muhammad-etal-2022-naijasenti}
Shamsuddeen~Hassan Muhammad, David~Ifeoluwa Adelani, Sebastian Ruder,
  Ibrahim~Sa{'}id Ahmad, Idris Abdulmumin, Bello~Shehu Bello, Monojit
  Choudhury, Chris~Chinenye Emezue, Saheed~Salahudeen Abdullahi, Anuoluwapo
  Aremu, Al{\'\i}pio Jorge, and Pavel Brazdil. 2022.
\newblock \href {https://aclanthology.org/2022.lrec-1.63} {{N}aija{S}enti: A
  {N}igerian {T}witter sentiment corpus for multilingual sentiment analysis}.
\newblock In \emph{Proceedings of the Thirteenth Language Resources and
  Evaluation Conference}, pages 590--602, Marseille, France. European Language
  Resources Association.

\bibitem[{Ndubuisi-Obi et~al.(2019)Ndubuisi-Obi, Ghosh, and
  Jurgens}]{ndubuisi2019}
Innocent Ndubuisi-Obi, Sayan Ghosh, and David Jurgens. 2019.
\newblock \href {https://doi.org/10.18653/v1/P19-1625} {Wetin dey with these
  comments? modeling sociolinguistic factors affecting code-switching behavior
  in nigerian online discussions}.
\newblock In \emph{Proceedings of the 57th Annual Meeting of the Association
  for Computational Linguistics}, pages 6204--6214, Florence, Italy.
  Association for Computational Linguistics.

\bibitem[{Och and Ney(2003)}]{och-ney-2003-systematic}
Franz~Josef Och and Hermann Ney. 2003.
\newblock \href {https://doi.org/10.1162/089120103321337421} {A systematic
  comparison of various statistical alignment models}.
\newblock \emph{Computational Linguistics}, 29(1):19--51.

\bibitem[{Ofulue and Esizimetor(2010)}]{ofulue2010}
Christine~I. Ofulue and David~O. Esizimetor. 2010.
\newblock Guide to standard naijá orthography. an nla harmonized writing
  system for naijá publications.
\newblock Accessed: 2022-07-28.

\bibitem[{Ogueji and Ahia(2019)}]{ogueji2019pidginunmt}
Kelechi Ogueji and Orevaoghene Ahia. 2019.
\newblock Pidginunmt: Unsupervised neural machine translation from west african
  pidgin to english.
\newblock In \emph{Proceedings of the Workshop on Machine Learning for the
  Developing World}. Advances in Neural Information Processing Systems.

\bibitem[{Ojarikre(2013)}]{ojarikre2013perspectives}
Anthony Ojarikre. 2013.
\newblock {Perspectives and problems of codifying Nigerian Pidgin English
  orthography}.
\newblock \emph{Perspectives}, 3(12).

\bibitem[{Oyewusi et~al.(2020)Oyewusi, Adekanmbi, and Akinsande}]{oyewusi2020}
Wuraola~Fisayo Oyewusi, Olubayo Adekanmbi, and Olalekan Akinsande. 2020.
\newblock {Semantic Enrichment of Nigerian Pidgin English for Contextual
  Sentiment Classification}.
\newblock In \emph{Proceedings of the Workshop on AfricaNLP}. International
  Conference on Learning Representations.

\bibitem[{Oyewusi et~al.(2021)Oyewusi, Adekanmbi, Okoh, Onuigwe, Salami,
  Osakuade, Ibejih, and Musa}]{oyewusi2021}
Wuraola~Fisayo Oyewusi, Olubayo Adekanmbi, Ifeoma Okoh, Vitus Onuigwe,
  Mary~Idera Salami, Opeyemi Osakuade, Sharon Ibejih, and Usman~Abdullahi Musa.
  2021.
\newblock Naijaner : Comprehensive named entity recognition for 5 nigerian
  languages.
\newblock In \emph{Proceedings of the Workshop for AfricaNLP}. Association for
  Computational Linguistics.

\bibitem[{Raffel et~al.(2020)Raffel, Shazeer, Roberts, Lee, Narang, Matena,
  Zhou, Li, and Liu}]{JMLR:v21:20-074}
Colin Raffel, Noam Shazeer, Adam Roberts, Katherine Lee, Sharan Narang, Michael
  Matena, Yanqi Zhou, Wei Li, and Peter~J. Liu. 2020.
\newblock \href {http://jmlr.org/papers/v21/20-074.html} {Exploring the limits
  of transfer learning with a unified text-to-text transformer}.
\newblock \emph{Journal of Machine Learning Research}, 21(140):1--67.

\end{thebibliography}

\label{lr:ref}
\bibliographylanguageresource{languageresource}



\appendix
\clearpage

\section{Construction of Orthographic Variation Types}
\label{appendix_variation_types}

\pinjie{To identify common orthographic variations, we first identified common differences between Pidgin and English spellings by aligning the Pidgin data with the English translations. We evaluated the variations of English words that occurred more than 100 times in the dataset. This resulted in a set of variation ``rules''. We then tested these rules by applying them to other English words (that is, generating more variations of other words) and manually annotating the generation quality (evaluated by the Pidgin speaker in the initial phase). This effort showed that some rules were not generalizable, i.e. they were only applicable to certain cases and potentially led to undesired variation generation. For instance, a rule for Pidgin `awa’ (English \textit{our}) is not generalizable to other words with “ou” (e.g. ‘sour’ or ‘flour’ would not become “sawa” or “flawa”). Such rules were eliminated and we kept the remaining rules that could be generalized to new variations (also see Table \ref{tab:variationclass}).}

\section{Generating irrelevant words
}
\label{appendix_irrelevant_words}

\pinjie{Our preliminary experiments showed an issue in generating variation candidates that match other existing words. For example, `deep’ could be synthesized as `dip’ by substituting `ee’ with `i’. However, such variations are unlikely to occur in the Pidgin datasets, since Nigerian Pidgin is an English-derived language and these syntheses conflict with an existing English word (note that it is not impossible to encounter such variations in natural language settings, but we do consider them to be noisy). We found that this issue could be easily resolved by filtering out generation candidates using the existing English lexicon. 
\\
Crucially, we found that a moderate level of ‘overgeneration’ could improve the generalization to unseen texts. 
The expanded training set with more diverse word forms enhances the model’s proficiency.}

\end{document}